  \pdfoutput=1
\documentclass{article}
\usepackage{ijcai19}
\usepackage{diagbox}
\usepackage{times}
\usepackage{soul}
\usepackage{url}
\usepackage[hidelinks]{hyperref}
\usepackage[utf8]{inputenc}
\usepackage[small]{caption}
\usepackage{graphicx}
\usepackage{amsmath}
\usepackage{amsfonts}
\usepackage{booktabs}
\usepackage{rotating}
\usepackage[protrusion=true,spacing=true]{microtype}
\microtypecontext{spacing=nonfrench}
\usepackage{float}
\usepackage[round]{natbib}
\title{Dealing with Non-Stationarity in Multi-Agent Deep  Reinforcement Learning}

\author{
Georgios Papoudakis\and
Filippos Christianos\and
Arrasy Rahman \and
Stefano V. Albrecht\\
\affiliations
School of Informatics, The University of Edinburgh \\
\emails
\{g.papoudakis, f.christianos, arrasy.rahman, s.albrecht\}@ed.ac.uk}

\begin{document}

\maketitle

\begin{abstract}
Recent developments in deep reinforcement learning are concerned with creating decision-making agents which can perform well in various complex domains. A particular approach which has received increasing attention is multi-agent reinforcement learning, in which multiple agents learn concurrently to coordinate their actions. In such multi-agent environments, additional learning problems arise due to the continually changing decision-making policies of agents. This paper surveys recent works that address the non-stationarity problem in multi-agent deep reinforcement learning. The surveyed methods range from modifications in the training procedure, such as centralized training, to learning representations of the opponent’s policy, meta-learning, communication, and decentralized learning. The survey concludes with a list of open problems and possible lines of future research.

\end{abstract}

\section{Introduction}
Deep learning has revolutionized the development of agents which can act autonomously in complex environments. Traditional reinforcement learning (RL) methods, using tabular representations or linear function approximators, are difficult to scale to high-dimensional environments, and are mostly applied to small grid worlds or environments that provide a high-level state representation. The combination of deep learning with existing RL algorithms enabled the development of agents that act in environments with larger state spaces, such as images. Recently, some promising deep RL algorithms that can handle large state environments both with continuous and discrete action spaces have been proposed  \citep{mnih2015human,schulman2017proximal,fujimoto2018addressing}.

%  \citet{mnih2015human} proposed a successful way to combine neural networks with Q-learning \citep{watkins1992q}.
% Despite the recent success of deep RL in single-agent environments, it does not
% fully address the requirements of multi-agent systems. Autonomous vehicles \citep{wurman2008coordinating},
% network packet routing \citep{boyan1994packet} and garbage collection \citep{makar2001hierarchical} are
% just a few examples of multi-agent systems. There are a plethora of reasons that deep RL is not compatible with multi-agent systems.  The major problem is the non-stationarity of multi-agent environments due to the fact that all the agents change their policy during the training procedure. We
% are going to analyze this issue in section \ref{sec:2}. There are various other issues that multi-agent systems
% face, like the credit assignment problem and the partial observability. At each timestep, only a small
% number of agents contribute to the reward. Therefore, we may reward agents that did not contribute
% to the solution, or punish agents that acted optimally. Additionally, in most multi-agent environments
% each agent has access only to its local observation and not in the true state of the environment. This
% significantly hurts the performance and stability of the training.

Multi-agent systems consist of multiple agents acting and learning in a shared environment. Many real-world decision making problems can be modeled as multi-agent systems, such as autonomous vehicles, resource allocation problems, robot swarms, and human-robot interaction. Despite the recent success of deep RL in single-agent environments, there are additional challenges in multi-agent RL.
% Autonomous vehicles \citep{wurman2008coordinating}, network packet routing \citep{boyan1994packet} and garbage collection \citep{makar2001hierarchical} are a few examples of multi-agent systems. 
 One major challenge, and the focus of this survey, is the non-stationarity of multi-agent environments created by agents that change their policies during the training procedure. This non-stationarity stems from breaking the Markov assumption that governs most single-agent RL algorithms. Since the transitions and rewards depend on actions of all agents, whose decision policies keep changing in the learning process, each agent can enter an endless cycle of adapting to other agents.
 
Additional problems in multi-agent systems include multi-agent credit assignment, partial observability, and heterogeneity. In the first problem, only a subset of agents contribute to a reward, and we need to identify and reward them while avoiding punishing agents that acted optimally. Partial observability consists of agents having access only to their local observations and not the actual state of the environment, which can significantly hinder training performance. Heterogeneity refers to the fact that agents may have different sensing and acting capabilities, and that their learning algorithms may differ \citep{ar2012}.

Multi-agent RL is a widely studied topic with surveys ranging from modelling other agents~\citep{albrecht2018autonomous} to transfer learning~\citep{da2019survey} and non-stationarity~\citep{hernandez2017survey}. However, there is a recent focus on multi-agent deep  RL which is not discussed in the above surveys. In this work, we aim to consolidate progress in the area, and discuss non-stationarity in deep multi-agent RL settings. The survey concludes by discussing open problems and possible future research directions.

% After this brief introduction to the topic that we are going to discuss, we will refer to the structure
% of this review: In section \ref{sec:2}, we will refer to background information in reinforcement
% learning. In section \ref{sec:3}, we will describe a large number of recent works on deep multi-agent RL, that address non-stationarity. Finally, in section \ref{sec:4} we will present some open problem and direction for future research, and we will conclude our survey in section \ref{sec:5}.
%After this brief introduction to the topic, we have to note that our survey is the only one that provides a unified view of recent approaches in dealing with non-stationarity  in deep multi-agent RL. For other relevant topics, such as opponent modelling, refer to other surveys such as the work of \cite{albrecht2018autonomous}. Additionally, \citet{hernandez2017survey} refer to the problem of non-stationarity by categorizing different methods. However, they do not focus on recent deep RL works.  

\section{Background} 
\label{sec:2}
In this section, we will briefly refer to definitions and previous works on RL, and we will formulate the non-stationarity problem of multi-agent RL.

\subsection{Markov Decision Processes}
We can mathematically model a decision making problem using Markov Decision Processes (MDP). An MDP is defined as $ (S, A, r, T) $, where $ S $ is the set of possible states, $A$ is the set of available actions, $ r: S \times A \times S \rightarrow \mathbb{R} $ is the reward function and $ T: S \times A \times S \rightarrow [0,1] $ is the probability distribution over next states given the current state and action. Additionally, in every MDP a stochastic policy function $\pi: S \times A \rightarrow [0,1]$ or a deterministic policy function $ \mu: S \rightarrow \mathbb{R}^{|A|} $ is defined to select an action for the current state. 

Given a policy $\pi$, an MDP has two value functions. The state value function $V:S \rightarrow \mathbb{R}$ is the expected sum of discounted rewards from a given state under the policy $\pi$;  $ V^{\pi}(s) = \mathbb{E}_{\pi}[\sum_{t=0}^{N}\gamma^t r_t | s] $, and the state-action value function $Q:S\times A \rightarrow \mathbb{R}$ is the expected sum of discounted rewards from a given state and action under the policy $\pi$;
$Q^{\pi}(s, a) = \mathbb{E}_{\pi} [ \sum_{t=0}^{N}\gamma^t r_t | s, a ] $.  In the equations $\gamma$ is the discount factor, taking values in $[0,1]$, and $r_t$ is the reward after $t$ time steps. 

% Using Bellman equations, we can also express the value functions in an iterative form. That is, an expression of the value function at the current timestep using the next timestep:
% $$V^{\pi}(s) = \sum_a \pi(a|s) \sum_{s'} T(s,a,s')[r(s,a,s') + \gamma V^{\pi}(s')]$$
% $$Q^{\pi}(s, a) =  \sum_{s'} T(s,a,s')[r(s,a,s') + \gamma \sum_{a'} \pi(a'|s')Q^{\pi}(s', a')]$$

Solving an MDP requires computing the optimal policy. That is the policy that maximizes the expected discounted sum of rewards $V^*(s) = \max_{\pi}V^{\pi}(s)$. If access to the reward and transition function is available, this problem can be solved by iterating the Bellman equations and using dynamic programming. However, these two functions are unknown in most scenarios. In this case, RL approaches are used to compute the optimal policy. 

\subsection{Reinforcement Learning Methods}
RL consists of multiple methods for solving MDPs. Temporal difference (TD) methods work by estimating the value functions (or the state-action value function) by interacting with the environment. The greedy policy is computed using the following expression: $\pi(s) = \arg\max_{a} Q(s,a)$. A commonly used method is Q-learning \citep{watkins1992q}. The state-action value function (Q-values) can be expressed in tabular form (storing a separate value for every state and action) or can be approximated using a parameterized function. It is then updated by minimizing the following loss function:
$L = \frac{1}{2}(r + \gamma \max_{a'}Q(s',a')- Q(s,a))^2$. This loss can be minimized using a gradient optimizer, either with respect to Q-values (in the tabular case) or the parameters. 
 
 An alternative model-free approach are policy gradient (PG) methods. Given a stochastic or a deterministic policy parameterized by parameters $\theta$, the goal is to compute the policy that maximizes the expected discounted sum of rewards from the first state. Therefore, the objective function is $J = V^{\pi}(s_0)$. Using the policy gradient theorem, the gradient of the objective with respect to the parameters of the policy can be computed by maximizing the objective using a gradient-based optimizer. In the case of stochastic policy \citep{sutton2000policy} $\nabla_{\theta}J = \mathbb{E}[G_t \nabla_{\theta} \log \pi_{\theta}(a|s)]$. In the case of deterministic policy \citep{silver2014deterministic}  $\nabla_{\theta}J = \mathbb{E}[\nabla_{\theta}\mu(s)\nabla_{a}Q(ts,a) |_{a=\mu(s)} ]$.

% The introduction of deep networks in RL algorithms greatly improved their ability to tackle problems. While the above algorithms have convergence guarantees i 
% restricts their possibilities
% to approximate Q values instead.
In the tabular case or when combined with linear approximators, both TD and PG methods are difficult to scale to large action and state spaces. Thus, deep networks have been used \citep{mnih2015human} to address this issue.  \citet{mnih2015human} proposed \textit{experience replay} and the use of \textit{target networks} in order to deal with instability issues that come with deep networks.

\subsection{Markov Games}

Markov games \citep{littman1994markov} are a generalization of MDP to multi-agent settings. A Markov game is defined as $(I, S, A, r, T)$ where, $I$ is the set of $N$ agents, $S$ is the set of states, $ A=A_1 \times A_2 ... \times A_{N} $ is the set of actions of all the agents, $r = (r_1, r_2, ..., r_{N})$, where $r_i: S \times A \times S \rightarrow \mathbb{R}$ is the reward function of agent $i$. Usually, in cooperative setting $r_1 = r_2 ... = r_ N$. $T: S \times A \times S\rightarrow [0, 1] $ is the probability distribution over next states, given the current state and the actions of all agents.

In this survey, we assume that each agent has access only to a local observation $o$ and not to the full state of the environment. As a result, the policy of the agents is conditioned on their history of observations $h_i =[o_{i,t}, o_{i,t-1}, o_{i,t-2}, ...]$. Given that the history of all agents $H = [h_1, h_2, ..., h_N]$, the joint policy is described as $\pi(\mathbf{a}|H) = (\pi_1(a_1|h_{1}), \pi_2(a_2|h_{2}), ..., \pi_N(a_N|h_{N}))$. A common simplification is to condition the policy of each agent's only on the most recent local observation,  $\pi(\mathbf{a}|o_t) = (\pi_1(a_1|o_{1,t}), \pi_2(a_2|o_{2,t}), ..., \pi_N(a_N|o_{N,t}))$. However, conditioning on the history of observations experimentally leads to better results in partially observable environment.
Similarly to MDP, the goal of each agent is to maximize its expected discounted sum of rewards $ V^{\pi_i}_i(h_i) = E [\sum_{i=0}\gamma^i r_i | h_i]$.

% each agent's observation,  $\pi(\mathbf{a}|s_t) = (\pi_1(a_1|o_{1,t}), \pi_2(a_2|o_{2,t}), ..., \pi_N(a_N|o_{N,t}))$. An incremental approach, that results in better performance, is to condition the policy of each agent on their history of observations $h_i =[o_{i,t}, o_{i,t-1}, o_{i,t-2}, ...]$ instead of a single observation. Given that the history of all agents $H = [h_1, h_2, ..., h_N]$ the joint policy is described as $\pi(\mathbf{a}|H) = (\pi_1(a_1|h_{1}), \pi_2(a_2|h_{2}), ..., \pi_N(a_N|h_{N}))$ 

\subsection{Centralized and Decentralized Architectures}

Two main architectures can be used for learning in multi-agent systems. The first architecture is centralized learning. In centralized learning, agents are jointly modelled and a common centralized policy for all the agents is trained. The input to the network is the concatenation of the observation of all the agents, and the output is the combination of all the actions. The most important issue of centralized architectures is the large input and output space. The input increases linearly and the output exponentially with respect to the number of agents.

On the other hand, in decentralized learning, the agents are trained independently from the others. Each agent has its policy network, takes a local observation as input and outputs an action. Although this method does not suffer from scalability issues, many different problems arise as well. Some of these issues are non-stationarity of the environment, the credit assignment problem and lack of explicit coordination. In this survey, we focus on works that attempt to handle issues related to non-stationarity.

\subsection{The Non-Stationarity Problem}

In Markov games, the state transition function $T$ and the reward function of each agent $r_i$ depend on the actions of all agents. During the training of multiple agents, the policy of each agent changes through time. As a result, each agents' perceived transition and reward functions change as well. Single-agent RL procedures which commonly assume stationarity of these functions might not quickly adapt to such changes.

%these functions unstable and inefficient. 

%F (RE2 - example)
As an example, consider a simple game like repeated Rock-Paper-Scissors. In it, two agents simultaneously perform actions \(a_i \in \{a_{paper}, a_{rock}, a_{scissors}\}\) and receive their respective rewards. The agents, seeking to maximize their reward, learn a policy \(\pi_i\) using the interaction history. Ideally, the policy would learn to play actions that beat the opponent's most used actions. For instance, if agent $1$ showed a preference to $a_{paper}$ then $\pi_2$ would learn to prioritize $a_{scissor}$. Subsequently, agent $2$ accumulates a history with that action, to which agent 1 reacts by changing its policy accordingly. This process may continue in this fashion, leading to what we refer to as non-stationarity due to changing behaviors.

\section{Dealing with Non-Stationarity}
\label{sec:3}
The following subsections survey various approaches that have been proposed to tackle non-stationarity in multi-agent deep RL. These approaches range from using modifications of standard RL training methods to computing and sharing additional opponent information. Details are summarized in Table~\ref{tab:1}.

\subsection{Centralized Critic Techniques}

A step toward dealing with non-stationarity is the centralized critic architecture. For this architecture, an actor-critic algorithm is used, which consists of two components. The critics' training is centralized and has access to the observations and actions of all agents, while the actors' training is decentralized. Since the actor computes the policy, the critic component can be removed during testing, and therefore the approach has fully decentralized execution. By having access to the observations and the actions of the opponent during training, the agents do not experience unexpected changes in the dynamics of the environment, which results in stabilization of the procedure.

\citet{foerster2018counterfactual} used the actor-critic algorithm with stochastic policies to train agents and evaluated their method in Starcraft. The authors proposed a single centralized critic for all the agents and a different actor for each agent. Additionally, they proposed a modification in the advantage estimation of the actor-critic $A(s,a) = Q(s,a) - \sum_{a'_i}\pi_i(a'_i|o_i)Q(s, (a_{-i}, a'_i))$. This modification serves two purposes. First of all, the advantage estimation, which is used in the policy gradient, is conditioned on the observations and actions of all the agents. Since each agent has access to the observations and actions of all the other agents, the policy gradient estimation is conditioned on the policy of the other agents and therefore the non-stationarity is addressed. Additionally, this counterfactual advantage can be interpreted as the value of action $a$ compared to all the other action values. As a result, this advantage might be utilized to address the credit assignment problem, and this is the core contribution of the paper.

\citet{lowe2017multi} proposed a multi-agent architecture using the deterministic policy gradient \citep{silver2014deterministic} algorithm (MADDPG). In this method, each agent uses a centralized critic and a decentralized actor. Since the training of each agent is conditioned on the observation and action of all the other agents, each agent perceives the environment as stationary.  Another extension of MADDPG is MiniMax MADDPG (M3DDPG) \citep{li2019robust}, which uses Minimax Q-learning \citep{littman1994markov} in the critic to exhibit robustness against different opponents with altered policies.

% Additionally, a benefit of MADDPG compared to COMA is that it does not suffer from huge action space. In the stochastic actor-critic of COMA, the output of the critic scales exponentially with respect to the number of agents. However, in the deterministic policy gradient architecture of MADDPG, the critic has one output and the actions of each agent are fed in the input. Therefore, the input to the critic scales linearly with the number of agents, leading to a more efficient algorithm.
\subsection{Decentralized Learning Techniques}

Handling non-stationarity in multi-agent systems does not necessarily require centralized training techniques. An alternative decentralized approach that has been explored to handle non-stationarity in multi-agent deep RL problems is self-play. This approach trains a neural network, using each agents' own observation as input, by playing it against its current or previous versions to learn policies that can generalize to any opponents. This approach can be traced back to the early days of TD-Gammon \citep{tesauro1995temporal} which managed to win against the human champion in Backgammon. More recently, self-play was extended to more complex domains such as Go \citep{silver2017mastering} and even complex  locomotion environments with continuous state and action space \citep{bansal2018emergent}.

In TD-Gammon, the neural networks are trained using temporal difference methods to predict the outcome of a game. Unlike recent approaches, self-play in TD-Gammon only plays against the current parameter setting of the neural network. It is noted in their work \citep{tesauro1995temporal} that using self-play in this way might result in a lack of exploration during training due to the neural networks always choosing the same sequence of actions in different episodes in self-play. In their case, this problem did not occur due to the dynamics in Backgammon being stochastic.

In more recent applications of self-play~\citep{silver2017mastering,bansal2018emergent}, an additional modification to the training process was made to ensure that the training is effective in environments with deterministic dynamics. In this case, recent self-play approaches stored the neural network parameters at different points during learning. Subsequently, the opponent during the self-play process is chosen by randomly choosing between the current and previous versions of neural network parameters. Apart from extending self-play to deterministic environments, this allowed the neural network to generalize against a broader range of opponents. As a result, self-play managed to train policies that can generalize well in environments like Go \citep{silver2017mastering} and even complex locomotion tasks \citep{bansal2018emergent}.

Another technique which has been used to allow decentralized training is by stabilizing experience replay. Despite playing an essential part in single-agent deep RL approaches \citep{mnih2015human}, due to environments' non-stationarity, experience replay might store experiences that are no longer relevant for decentralized learning which results in worse performance. \citet{foerster2017stabilising} proposed importance sampling corrections to adjust the weight of previous experience to current environment dynamics. When combined with independent Q-learning \citep{tan1993multi}, this resulted in significantly better performance in a specific task in the Starcraft game. 
\subsection{Opponent Modelling}

Another feasible direction that deals with non-stationarity is opponent modelling. By modelling the intentions and policies of other agents, the training process of the agents might be stabilized. Modelling other agents in multi-agent systems has been widely studied and offers many research opportunities~\citep{albrecht2018autonomous}. In this survey, we mostly focus on recent methods that learn models of opponents or use them to condition the agents' policy on them.

% \cite{Raileanu2018} suggest an approach where agents use their own policy to predict the behavior of others. This method, namely \textit{Self Other-Modelling} (SOM) employs an actor-critic architecture and reuses exactly the same network for estimating the goals of the other agents. In more detail, a network $f(s_{s/o}, z_s, \bar{z}_o)$, with the parameters being the state, the goal of the agent and the goal of the other agent respectively, is being used in a forward pass to decide on an action. However, the same network is also used by switching the order of $z_s$ and $\bar{z}_o$ to infer the goal of the other agent. Observing the actual actions of the opponent allows then the agent to back-propagate and optimize the vector of trainable parameters $\bar{z}_o$. The authors denote the network that is used to infer the goal of the opponent agent as $f_{other}$, even though it uses the same parameters. The network itself can be trained by any reinforcement learning algorithm (e.g. \textit{asynchronous advantage actor-critic}~\cite{mnih2016asynchronous}), while $\bar{z}_o$ is trained by minimizing the cross-entropy between the real action of the agent and the policy that is produced by the $f_{other}$ network. While this method is more expensive, requiring online back-propagation at each step, it can be attractive due to its simplicity and the fact that it uses only one set of parameters for modelling others.
\citet{Raileanu2018} suggested an approach where agents use their policy to predict the behaviour of other agents. This method employs an actor-critic architecture and reuses the same network for estimating the goals of the other agents. In more details, a network $f(s_{s/o}, z_s, \bar{z}_o)$, with the inputs being the state, the goal of the agent and the goal of the other agent respectively, is being used in a forward pass to decide on an action. However, the same network is also used by switching the order of $z_s$ and $\bar{z}_o$ to infer the goal of the other agent. Observing the actual actions of the opponent allows the agent to back-propagate and optimize the vector of trainable parameters $\bar{z}_o$. 
% While this method is more expensive, requiring online back-propagation at each step, it can be attractive due to its simplicity and the fact that it uses only one set of parameters for modelling others.
In contrast, \citet{hehe2016opponent} developed a second, separate network to encode the opponent's behaviour. The combination of the two networks is done either by concatenating their hidden states or by the use of a mixture of experts. This separate network enables faster learning and even allows modelling of changing opponent behavior.

\citet{zhang2010multi} and \citet{foerster2018learning}  proposed a modification of the optimization function in policy gradient methods to incorporate the learning procedure of the opponents in the training of their agent. Given two agents with parameters  $\theta_1$ and $\theta_2$ respectively, the authors proposed the optimization of $V_1(\theta_1, \theta_2 + \Delta \theta_2)$, where $\Delta \theta_2 = \nabla_{\theta_{2}} V_2 ( \theta_1, \theta_2 )$ instead of the standard $V_1(\theta_1, \theta_2 )$. In this way, the training agent has access to the learning trajectory of the opponents, and therefore, the training procedure does not suffer from non-stationarity. \citet{zhang2010multi} assumed that the term $\Delta \theta_2$ is not differentiable with respect to $\theta_1$ and they proved convergence to Nash equilibrium in 2-player 2-action games. On the other hand, \citet{foerster2018learning} proposed \textit{Learning with Opponent Learning Awareness} (LOLA) where the term $\Delta \theta_2$ is differentiable with respect to $\theta_1$ to exploit the opponent learning dynamics. LOLA experimentally led to tit-for-tat behaviour in a different number of games and successfully managed to cooperate in the Independent Prisoners Dilemma (IPD). Most works, such as \citet{bowling2001rational} deviate in the IPD settings.

In order to be able to keep the stability of \citet{zhang2010multi} and the opponent dynamics exploitation of  \citet{foerster2018learning}, \citet{letcher2018stable} proposed \text{Stable Opponent Shaping} (SOS). \citet{letcher2018stable} provided examples of differentiable games, where Stable Fixed Points exhibit better behaviour than Nash equilibrium and examples where LOLA fails to converge in an Stable Fixed Point. For this reason, the authors proposed a \textit{partial stop-gradient operator}, which controls the trade-off between the two methods. SOS has guarantees of convergence, while at the same time, it results in the same or better performance than LOLA.

% ~\cite{foerster2018learning} proposed the \textit{Learning with Opponent Learning Awareness} (LOLA) method. Assuming that our agent has parameter $\theta_1$ and the opponent has parameters $\theta_2$, the authors proposed to optimize, using a policy gradient method, $V_1(\theta_1, \theta_2 + \Delta \theta_2)$, where $\Delta \theta_2 = \nabla_{\theta_{2}} V_2 ( \theta_1, \theta_2 )$ is the learning step of the opponent. In this way, the training agent has access to the learning trajectory of the opponents and therefore, the training procedure does not suffer from non-stationarity. LOLA assumes that the term $\Delta \theta_2 $ is differentiable respect to the parameters $\theta_1$

Another facet of opponent modelling that was enabled by the recent advances in training neural networks is related to learning representations for multi-agent learning. Approaches that fall under this category learn the representations by imposing a certain model structure to compute the representations. Architectures such as graph neural networks \citep{tacchetti2018relational}, feed-forward neural networks \citep{grover2018learning}, and recurrent neural networks \citep{rabinowitz2018machine} can be used to produce the representations.

Specific loss functions are used along with gradient-based optimization to train these models to output representations which can predict specific information of the opponents such as their actions \citep{rabinowitz2018machine,grover2018learning} or returns \citep{tacchetti2018relational} received by the modelled agent. The representation models can be trained using loss functions commonly used in supervised learning. Additionally, \citet{grover2018learning} provided an example loss function that combines action prediction loss while also maximizing the difference between representations of different agents' policies.

The representation networks are provided with opponents' observations as input during learning. The policy networks are then trained by receiving agent observations which have been augmented with output representations from the representation networks as input. Under the assumption that these trained representation models could generalize to opponents that have yet been encountered, these models should be able to provide additional information which might characterize opponents' policy. Empirically, this produced an increased performance against either learning \citep{rabinowitz2018machine} or stationary opponents in various learning environments.

\subsection{Meta-Learning}
%As an alternative to approaches that augment the learning algorithm with additional opponent information, meta-learning approaches can also handle non-stationarity. Meta-learning approaches assume that environment dynamics remain stationary for a short period of time before then changing. It then aims to find model and learning parameters that might adapt to this change of environment dynamics with the least amount of data.

Before advances in deep RL, approaches like tracking \citep{sutton2007role} and context detection \citep{da2006dealing} were proposed to allow quicker adaptation in non-stationary environments. Both approaches adopt a more reactive view on handling non-stationarity by using learning approaches that attempt to quickly change the policy or environment models once changes in environment dynamics occur. However, the results of locomotion tasks proposed by \citet{finn2017model} highlighted how reactive approaches such as tracking still cannot produce a quick adaptation of deep RL algorithms to changing dynamics using only a few learning updates. 

%As an example,~\cite{finn2017model} experimented on a single-agent locomotion domain where an agent is required to move and then achieve and maintain a certain velocity throughout the episode. In their experiment, the target velocity is changed between episodes and agents are required to quickly learn a new policy to adapt to it. Their result demonstrated that even a neural network pre-trained to achieve this goal on different sets of velocities could not learn to adapt in only a few learning updates. 

Instead of formulating a learning algorithm which can train deep neural networks to react to changing environment dynamics, another approach is to anticipate the changing dynamics. An optimization process can then be formulated to find initial neural network parameters that, given the changing dynamics, can learn using small amounts of learning updates. Meta-learning approaches like \textit{Model Agnostic Meta Learning} (MAML) \citep{finn2017model} specifically addresses optimization for this particular problem. \citet{al2017continuous} further extended MAML to handle non-stationarity in multi-agent problems.

%In \citet{al2017continuous}, adaptation towards non-stationarity was tested on \textit{iterated adaptation games} where an agent repeatedly play against the same opponent with both only allowed to learn in between each game. Their approach treated different game episodes as separate tasks. MAML was then used to train agents against a predefined collection of opponents to find initial model parameters which can quickly adapt to changes in dynamics using usual policy gradient updates. 

The proposed method by \citet{al2017continuous} provided an optimization process to search for initial neural network parameters, $\theta$, which can quickly adapt to non-stationarity. The optimization process first generates rollout data,~$\tau_{\mathcal{T}_{i}}$, from task $\mathcal{T}_{i}$. It then updates $\theta$ according to a policy gradient update rule to produce updated parameters $\phi$. The method subsequently runs the policy parameterized by$~\phi$ in the following task, $\mathcal{T}_{i+1}$, which results in a performance denoted by~$\mathcal{R}_{i+1}(\phi)$. The proposed approach finally utilizes a policy gradient algorithm to search for $\theta$ which maximizes the expected performance after update,~$\mathbb{E}_{\tau_{\mathcal{T}_{i}}\sim P_{T}(\tau|\theta)}\left[\mathbb{E}_{\tau_{\mathcal{T}_{i+1}}\sim P_{T+1}(\tau|\phi)}\left[\mathcal{R}_{i+1}(\phi)|\tau_{\mathcal{T}_{i}},\theta\right]\right]$. By explicitly optimizing the initial model parameters based on their expected performance after learning, the proposed meta-learning approach was able to significantly outperform adaptation methods such as tracking and other meta-learning adaptation strategies which have performed well in single-agent environments. This was tested in iterated adaptation games where an agent repeatedly play against the same opponent while only allowed to learn in between each game. 
%The usage of pre-trained neural networks have resulted in remarkable results in training supervised learning approaches to adapt to new tasks using a limited amount of data. However, this technique proved to be rather ineffective in reinforcement learning. This raises questions on whether relying on the task distributions during pre-training is the best solution for pretraining in reinforcement learning. 

%Instead of using approaches that reacts to changes in environment dynamics, meta-learning anticipates the possible changes which might happen during training and searches for policies which can quickly perform well using a few learning updates. 

%Few shot learning

\subsection{Communication}
Finally, the last category of methods that we discuss for handling non-stationarity is communication. Through communication, the different training agents can exchange information about their observations, actions and intentions to stabilize their training. While communication in multi-agent systems is a well-studied topic, we will focus on recent methods that learn to communicate using multi-agent deep RL.

A step in this direction is the work of \citet{foerster2016learning1}, who proposed the \textit{Deep Distributed Recurrent Q-Networks}, an architecture where all the agents share the same hidden layers and learn to communicate to solve riddles. In the same direction, \citet{sukhbaatar2016learning} proposed \textit{CommNet}, an architecture, where the input to each hidden layer $h_j^i = f(h_{j-1}^i, c_{j-1}^i)$ for each agent $i$, is the previous layer $h_{j-1}^i$ and a communication message $c_{j-1}^i = \frac{1}{I - 1}\sum_{i'}h_{j-1}^i$, which is the average of the previous hidden layers of all the other agents. Therefore, the agents learn to communicate by sharing the extracted features of their observations in some cooperative scenarios. \citet{singh2018learning} proposed  the \textit{Individualized Controlled Continuous Communication Model} (IC3NET), which is an extension of CommNet in competitive setting. In IC3NET, there is an additional communication gate, which either allows or blocks communication between agents, for example in competitive settings. 

All of the previous approaches assume that all agents have access to the hidden layers of the other agents. An elimination of this constraint was proposed by \citet{foerster2016learning}. The authors first suggested the \textit{Reinforced Inter-Agent Learning}, where each agent has two Q-networks. The first network outputs an action and the second a communication message which is fed in the Q-networks of the other agents. Both networks are trained using DQN. In the same work, they also developed the  \textit{differentiable inter-agent learning}, where they train only the action network with DQN, while they push the gradients of the other agents, through the communication channel to the communication network. This approach is similar to the work of \citet{mordatch2018emergence}, where the authors proposed a model that takes as input the messages of other agents and learns to output an action and a new communication message.

\section{Open Problems}
\label{sec:4}
%After describing a large number of works for addressing non-stationarity in multi-agent RL, in this section, we will refer to a number of open problem that are worth investigating from the community.
Based on the methods outlined in this survey, we identify several open problems with regards to non-stationarity and possible lines of future research.

\subsection{Transfer Learning for Non-Stationarity}

%Since the inherent non-stationarity in multi-agent systems is caused by learning opponents, one might assume that the environment dynamics will remain the same when no agent learns and update their policies. During this time where none of the opponents learns, the environment will appear stationary to the learning agent. Hence, non-stationarity in this case might be viewed as a sequence of shorter interactions with stationary dynamics until one of the opponents changes their policies.

%An interesting question related to the possibility of leveraging knowledge from past interactions under different environment dynamics would then arise as a result of this view. Under the multi-agent with learning opponents setup, agents need to formulate approaches which can decide which interactions in the past are relevant to the current interaction and quickly change their policies in very few updates using this knowledge. Therefore, this is similar to problems in the transfer learning setup.

Several transfer learning approaches for multi-agent systems have been covered in this survey. In this case, the learned representations and initialization values from meta-learning~\citep{al2017continuous} and approaches which learn opponent representations~\citep{grover2018learning, tacchetti2018relational, rabinowitz2018machine} can be viewed as transferred knowledge which might result in quicker adaptation to non-stationarity. Despite recent advances, there are still open questions regarding the form of the knowledge being transferred and how to leverage them for quicker adaptation.

\subsection{Open Multi-Agent Systems}
In real-world problems, the number of agents in an environment can be large and diverse. Furthermore, the number of agents might change due to agents leaving or entering the environment. This problem setting is often termed an \emph{open} multi-agent system~\citep{chandrasekaran2016individual}. The change in the number of agents can cause an action to have different consequences at different points during learning. For example, a certain action might lead to situations with high returns when another cooperative agent is in the environment, while also being inconsequential when the other agent left the environment.

None of the techniques presented in this survey were tested in environments with changing number of agents. In general,
transfer learning approaches that can reuse knowledge between problems with a varying number of agents might be a potential avenue of research in this topic. Furthermore, research on how agents can effectively deal with heterogeneity in capabilities and learning algorithms \citep{ar2012} will play an important part.
%\begin{itemize}
%\item By letting policy networks learn how to use the representations. But is it easy to generalize to these current representations?
%\item Does it generalize well, e.g. in MAML, how will it fare against opponents that have yet been encountered?
%\end{itemize}

\subsection{Limited Access to Opponent Information}

A large number of the works that we presented in opponent modelling, such as \citep{hehe2016opponent,grover2018learning}, requires access to the opponent's observations and chosen actions. While this is not a strong assumption during centralized training, it is very limiting during testing, especially when there is not established communication between the agents. More precisely, assuming that we have access in the observations and actions of the opponents during testing is too strong. Therefore, it is an open problem to create models that do not rely on this assumption.

\subsection{Convergence Properties}

An open problem with current multi-agent deep RL methods is the lack of theoretical understanding of their convergence properties and what types of outcomes they tend to achieve. A theoretical concept that could be used to encourage convergence are game-theoretic equilibria, such as Correlated and Nash equilibrium. One drawback of these approaches is the requirement of computing equilibrium solutions \citep{greenwald2003correlated}, as well as non-uniqueness of equilibria which requires some form of coordinated equilibrium selection. A recent example in this direction is the work by \citet{li2019robust} who used an approximate solution of the Minimax equilibrium. Therefore, an interesting research direction is to investigate approximate solutions in order to incorporate the notion of these equilibria in multi-agent deep RL.

\subsection{Multi-Agent Credit Assignment}
Another problem which arises from approaches with decentralized execution is related to assigning credit to agents in the environment. Despite improving centralized training~\citep{foerster2018counterfactual} or learning opponent representations~\citep{tacchetti2018relational}, methods to handle this problem can still be improved. In general, finding alternative neural network architectures and learning approaches that can decompose rewards~\citep{rashid2018qmix, sunehag2018value} for smaller group of agents can be a possible future research direction.

\section{Conclusion}

\label{sec:5}

To summarize our contribution, we identified five different approaches for handling non-stationarity in  multi-agent deep RL. In order to understand their characteristics and their differences, we provided a detailed categorization in Table \ref{tab:1}. Finally, we outlined several open problems with respect to non-stationarity and possible directions of future research.

% We summarize the works that we presented in Table \ref{tab:1}. We categorize these works based on their training and execution architecture, their modelling technique, whether it is applied in a cooperative, competitive or mixed settings, the algorithm which is used for RL training and the number of opponents that they can handle. We have outlined how these approaches have certain shortcomings and highlighted potential interesting future research directions.

%\section*{Acknowledgement}

\begin{table*}[h!]

  \centering
  \begin{tabular}{l c c c c c c c}
     &
     \rotatebox{75}{Settings} & \rotatebox{75}{Training} & \rotatebox{75}{Execution} & \rotatebox{75}{Modelling}  & \rotatebox{75}{Opp. Info} &  \rotatebox{75}{Algorithm} & \rotatebox{75}{Num agents} \\ 
    \hline \hline
    \citet{tacchetti2018relational} & Mixed   & Centr. &  Decentr. & Explicit & Obs / actions & A2C  & $\geq2$ \\
   \hline 
       \citet{singh2018learning} & Mixed & Decentr. & Decentr. & No & None & PG & $\geq2$ \\ \hline
        \cite{letcher2018stable} & Mixed & Decentr. & Decentr. &  Explicit & Parameters & PG  & $2$ \\
    \hline
    \citet{li2019robust} & Mixed & Centr. & Decentr. & No & Obs / actions  & DDPG & $\geq2$ \\
    \hline
       \citet{al2017continuous} & Comp. & Decentr. & Decentr. & No  & None  & PPO & $2$ \\
   \hline

     \citet{bansal2018emergent} & Comp.  & Decentr.. &  Decentr. & No & None & PPO  & $2$ \\
   \hline

    \citet{Raileanu2018} &  Mixed  & Centr. & Centr. & Explicit & Obs / actions & A3C  & $2$ \\
    \hline
    \citet{mordatch2018emergence} & Coop. & Decentr. & Decentr. & No & None & PG  &$\geq2$ \\ \hline
    \cite{foerster2018learning} & Mixed  & Decentr. & Decentr. &  Explicit & Parameters & PG & $2$ \\
    \hline
 
    \cite{grover2018learning} & Mixed  & Centr. & Centr. &  Explicit & Obs / actions& PPO / DDPG  & $2$ \\
    \hline
   \cite{rabinowitz2018machine} & Mixed  & Centr. & Centr. &  Explicit & Obs / actions & Imit.  & $\geq2$ \\ \hline
   \citet{foerster2018counterfactual} & Coop.  & Centr. & Decentr. & No  & Obs / actions & Actor-critic & $\geq2$ \\ \hline
   
   \citet{lowe2017multi} & Mixed & Centr. & Decentr. & No  & Obs / actions & DDPG & $\geq2$ \\
    \hline
       \cite{foerster2017stabilising} & Mixed & Decentr. & Decentr. & No & None & Q-learning & $\geq 2$ \\ 
   \hline
    \citet{sukhbaatar2016learning} & Coop. & Decentr. & Decentr. & No & None & PG & $\geq 2$ \\ \hline
   \citet{foerster2016learning} & Coop. & Centr. & Decentr. & No  & None & Q-learning & $\geq2$ \\ \hline
       \cite{hehe2016opponent} & Mixed  & Centr. & Centr. & Implicit & Obs  & Q-learning & $2$ \\
    \hline
        \cite{zhang2010multi} & Mixed & Decentr. & Decentr. &  Explicit & Parameters  & PG & $2$ \\
    \hline

  \end{tabular}
  \caption{Categorization of the surveyed algorithms that deal with non-stationarity. The algorithms are categorized based on the environment setting (cooperative, competitive, mixed), their training and execution method (centralized, decentralized), their type of modelling, the opponent information that they require, the learning algorithm that they use, and the number of agents that can be handled.}
  \label{tab:1}
\end{table*}

% \small
\newpage
\bibliographystyle{named}
\bibliography{ijcai19}
\end{document}